%% file: holo_fusion.tex
\documentclass[10pt,twocolumn,letterpaper]{article}
\usepackage{styles/iccv}

\usepackage{algorithm}
\usepackage{algpseudocode}
\usepackage{amsmath}
\usepackage{amssymb}
\usepackage{booktabs}
\usepackage{caption}
\usepackage{colortbl}
\usepackage{graphicx}
\usepackage{multirow}
\usepackage{pifont}
\usepackage{siunitx}
\usepackage{tabularx}
\usepackage{times}
\usepackage{xcolor}
\usepackage{xspace}
\usepackage[pagebackref=true,breaklinks=true,letterpaper=true,colorlinks,bookmarks=false]{hyperref}
\usepackage[capitalize]{cleveref}

\crefname{section}{Sec.}{Secs.}
\Crefname{section}{Section}{Sections}
\Crefname{table}{Table}{Tables}
\crefname{table}{Tab.}{Tabs.}

\definecolor{maroon}{HTML}{efa884}
\definecolor{dorange}{HTML}{3b2000}
\definecolor{yelgreen}{HTML}{d2b946}
\hypersetup{citecolor=yelgreen}
\hypersetup{linkcolor=yelgreen}
\hypersetup{linkcolor=yelgreen}
\hypersetup{urlcolor=yelgreen}

\makeatletter
\renewcommand{\paragraph}{%
  \@startsection{paragraph}{4}%
  {\z@}{0.25em}{-1em}%
  {\normalfont\normalsize\bfseries}%
}
\makeatother

\iccvfinalcopy
\ificcvfinal\fi

\input{includes/general_commands}

\newcommand{\name}{\textsc{HoloFusion}\xspace}

\title{\name: Towards Photo-realistic 3D Generative Modeling}

\author{Animesh Karnewar\\
UCL\\
{\tt\small a.karnewar@ucl.ac.uk}
\and
Niloy J. Mitra\\
UCL\\
{\tt\small n.mitra@ucl.ac.uk}
\and
Andrea Vedaldi\\
Meta AI\\
{\tt\small vedaldi@meta.com}
\and
David Novotny\\
Meta AI\\
{\tt\small dnovotny@meta.com}
}

\begin{document}
\ificcvfinal\thispagestyle{empty}\fi

\twocolumn[{
\renewcommand\twocolumn[1][]{#1}%
\maketitle
\thispagestyle{empty}
\vspace{-1.0cm}
\begin{center}
\centering
\captionsetup{type=figure}
\includegraphics[width=1.01\linewidth]{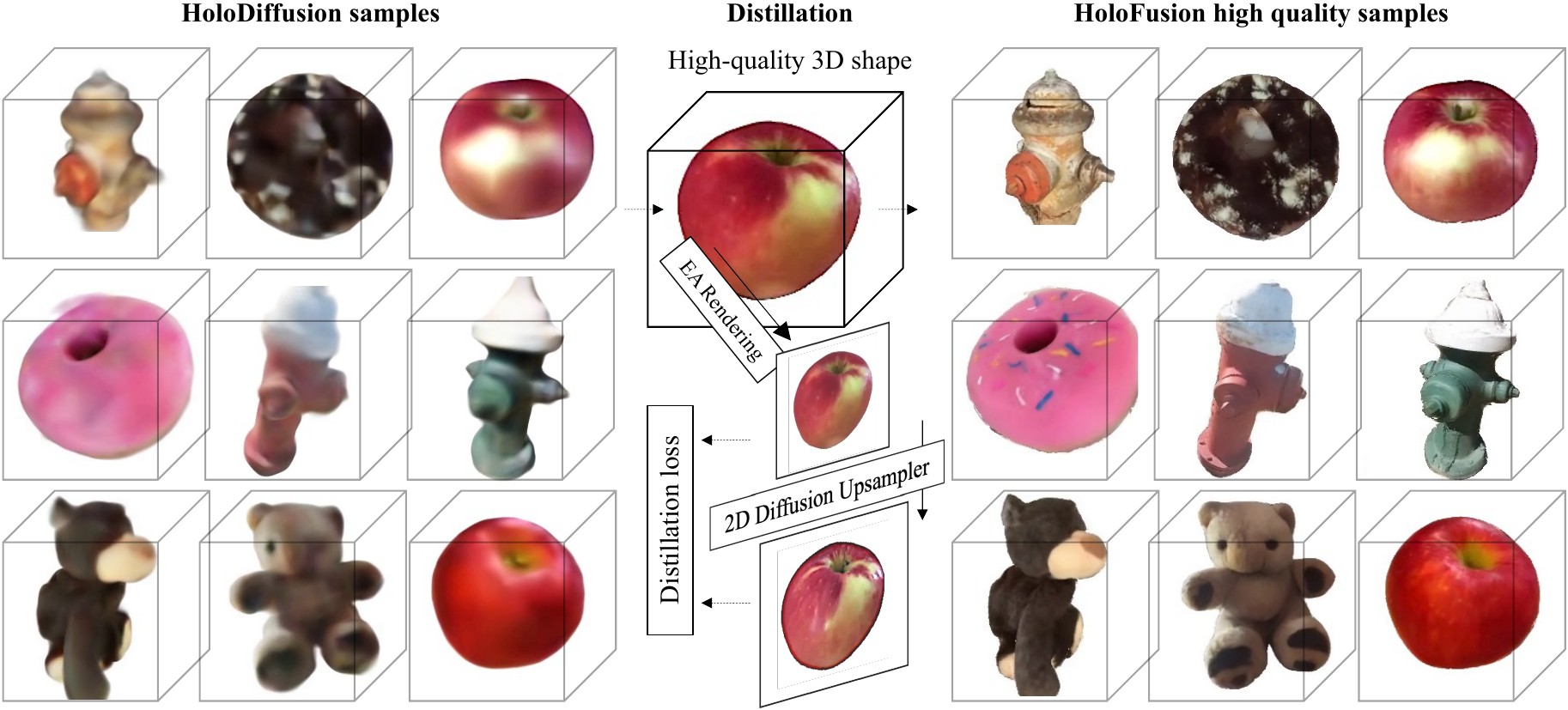}
\vspace{-0.5cm}
\captionof{figure}{We propose \name to generate photo-realistic 3D radiance fields by extending the HoloDiffusion method with a jointly trained 2D `super resolution' network. The independently super-resolved images are fused back into the 3D representation to improve the fidelity of the 3D model via distillation, while preserving the consistency across view changes.}%
\label{fig:teaser-fig}
\end{center}
}]

\input{src/abstract}
\input{src/intro}
\input{src/related}
\input{src/method}

\input{src/experiments}

\input{src/outro}
\input{src/acknowledgements}

{\small\bibliographystyle{bib/ieee_fullname}
\bibliography{bib/holo_fusion, bib/vedaldi_general, bib/vedaldi_specific, bib/holo_diffusion}}
\end{document}

%% file: includes/general_commands.tex
\definecolor{rowblue}{RGB}{220,230,240}
\definecolor{myorchid}{RGB}{150,10,30}
\definecolor{myblue}{RGB}{10,30,250}
\definecolor{mygreen}{RGB}{10,120,10}

\newcommand{\cmark}{\ding{51}}
\newcommand{\xmark}{\ding{55}}

\DeclareGraphicsExtensions{.jpg,.pdf,.png}

\definecolor{colorA}{HTML}{aa6600}
\definecolor{colorB}{HTML}{ff9912}
\definecolor{colorC}{HTML}{ffc125}
\definecolor{colorD}{HTML}{7a83e4}
\definecolor{colorE}{HTML}{cdc673}
\definecolor{colorF}{HTML}{cfb53b}
\definecolor{colorG}{HTML}{98a148}
\definecolor{colorH}{HTML}{666666}
\definecolor{colorI}{HTML}{ef5998}
\definecolor{colorJ}{HTML}{9112ff}

\definecolor{colorY}{HTML}{000000}
\definecolor{colorZ}{HTML}{777777}

\colorlet{color$\pi$-GAN}{colorA}
\colorlet{colorEG3D}{colorB}
\colorlet{colorGET3D}{colorC}
\colorlet{colorStable-DreamFusion}{colorD}
\colorlet{colorDreamFusion}{colorD}
\colorlet{colorHoloDiffusion}{colorE}
\colorlet{colorHoloDiffusion$^*$}{colorF}
\colorlet{colorHoloFusion}{colorG}
\colorlet{colorHoloFusion (Ours)}{colorG}
\colorlet{colorOurs}{colorG}
\colorlet{colorOur}{colorG}
\colorlet{colorHoloFusion (MSE)}{colorY}
\colorlet{colorMSE}{colorY}
\colorlet{colorHoloFusion (SDS)}{colorY}
\colorlet{colorSDS}{colorY}
\colorlet{colorHoloFusion (w/o Patch remix)}{colorY}

\newcommand{\method}[1]{\textcolor{color#1}{#1}}
\newcommand{\methodb}[1]{\textcolor{color#1}{\textbf{#1}}}

%% file: src/abstract.tex
\begin{abstract}
Diffusion-based image generators can now produce high-quality and diverse samples, but their success has yet to fully translate to 3D generation:
existing diffusion methods can either generate low-resolution but 3D consistent outputs, or detailed 2D views of 3D objects but with potential structural defects and lacking view consistency or realism.
We present \name, a method that combines the best of these approaches to produce high-fidelity, plausible, and diverse 3D samples while learning from a collection of multi-view 2D images only.
The method first generates coarse 3D samples using a variant of the recently proposed HoloDiffusion generator.
Then, it independently renders and upsamples a large number of views of the coarse 3D model, super-resolves them to add detail, and distills those into a single, high-fidelity implicit 3D representation, which also ensures view-consistency of the final renders.
The super-resolution network is trained as an integral part of \name, end-to-end, and the final distillation uses a new sampling scheme to capture the space of super-resolved signals.
We compare our method against existing baselines, including DreamFusion, Get3D, EG3D, and HoloDiffusion, and achieve, to the best of our knowledge, the most realistic results on the challenging CO3Dv2 dataset.
\vspace{-0.5cm}
\end{abstract}

%% file: src/intro.tex
\section{Introduction}%
\label{s:intro}

Diffusion models~\cite{saharia2022photorealistic,dhariwal2021diffusion,rombach2021stablediffusion} are at the basis of state-of-the-art 2D image generators which can now produce very high-quality and diverse outputs.
However, their success has yet to be translated to 3D and there is no generator that can produce 3D assets of a comparable quality.

Recent attempts at extending diffusion to 3D generation have reported mixed success.
Some authors have attempted to apply diffusion directly in 3D~\cite{karnewar2023holodiffusion}, or still in 2D but using a 3D-aware neural network~\cite{watson22novel,anciukevicius22renderdiffusion:}.
This requires solving two problems:
first, finding a suitable 3D representation (\eg, triplane features~\cite{chan2022efficient}, mesh~\cite{lin22magic3d:}, voxels~\cite{karnewar2023holodiffusion}) that scales well with resolution and is amenable to diffusion;
and, second, obtaining a large amount of 3D training data, for example using synthetic models~\cite{wang2022rodin,muller22diffrf:}, or training the model using only 2D images~\cite{karnewar2023holodiffusion}, often via differentiable (volume) rendering~\cite{henzler2019platonic-gan,mildenhall20nerf:}.
However, the quality of results so far is limited, especially when training on real images.

Other authors have proposed to \textit{distill} 3D objects from pre-trained 2D image generators.
For instance Score Distillation Sampling (SDS)~\cite{poole2022dreamfusion} can sample 3D objects from a high-quality off-the-shelf 2D diffusion model while requiring no (re){}training.
However, without any 3D guidance, distillation methods often produce implausible results;
for example, they suffer from the `Janus effect', where details of the front of the object are replicated on its back.
They also create overly-smooth outputs that average out inconsistencies arising from the fact that the signal obtained from
the 2D model is analogous to sampling independent views of the object (see \cref{ssec:quant_and_qual} for examples).
Furthermore, distillation methods do not support unconditional sampling, even if the underlying image generator does, as strong language guidance is required to stabilise the 3D reconstruction.

In this work, we propose \name, a method that combines the best of both approaches.
We start from HoloDiffusion~\cite{karnewar2023holodiffusion}, a diffusion-based 3D generator.
This model can be trained using only a multiview image dataset like~\cite{reizenstein21common} and produces outputs that are 3D consistent.
However, the output resolution is limited by computation and memory.
We augment the base model with a lightweight super-resolution network that upscales the initial renders.
Crucially, the 2D super-resolution model is integrated and trained jointly with the 3D generator, end-to-end.

The super-resolution network outputs detailed views of the 3D object, and the underlying 3D generator ensures that the coarse structure of these views is indeed consistent (\eg, avoiding the Janus effect and other structural artifacts).
However, the 2D upscaling still progresses independently for different views, which means that fine grained details may still be inconsistent between views.
We address this issue by distilling a single, coherent, high quality 3D model of the object from the output of the upsampler.
For this, we propose a new distillation technique that efficiently combines several putative super-resolved views of the object into a single, coherent 3D reconstruction.

With this, we are able to train a high-quality 3D generator model purely from real 2D data.
This model is capable of generating consistent and detailed 3D objects, which in turn result in view-consistent renderings (see \cref{fig:teaser-fig}) at a quality not achievable by prior methods.


We evaluate \name on real images (CO3Dv2 dataset~\cite{reizenstein21common}) and compare with a variety of competing alternatives (\eg, HoloDiffusion~\cite{karnewar2023holodiffusion}, Get3D~\cite{gao2022get3d}, EG3D~\cite{chan2022efficient}, DreamFusion~\cite{stable-dreamfusion}) demonstrating  that view-consistent high-quality 3D generation is possible using our simple, effective, easy-to-implement hybrid approach.

%% file: src/related.tex
\section{Related Work}%
\label{s:related}





\paragraph{3D generators that use adversarial learning.}


Generative Adversarial Learning (GAN)~\cite{goodfellow2020gan} learns a generator network so that its ``fake'' samples cannot be distinguished from real images by a second discriminator network.
Approaches such as PlatonicGAN~\cite{henzler2019platonic-gan}, HoloGAN~\cite{nguyen-phuoc19hologan:}, and PrGAN~\cite{gadelha163d-shape} introduced 3D structure into the generator network, achieving 3D shape generation with only image-level supervision. Our method is related to those as it renders images from a  generated voxel grid, as well as to HoloGAN~\cite{nguyen-phuoc19hologan:}, which renders features and then converts them into an image by a lightweight 2D convolutional network.
Other voxel-based 3D generators include VoxGRAF~\cite{schwarz2022voxgraf} and NeuralVolumes~\cite{lombardi2019neural}.

More recently, 3D generators have built on neural radiance fields~\cite{mildenhall20nerf:}.
GRAF~\cite{schwarz20graf:} was the first to adopt the NeRF framework;
analogous to PlatonicGAN, they generate the parameters of an MLP which renders realistic images of the object from a random viewpoint.
This idea has been improved in StyleNeRF~\cite{gu21stylenerf} and EG3D~\cite{chan2022efficient} by adding a 2D convolutional post-processing step after emission-absorption rendering, which is analogous to our super-resolution network.
EG3D also introduced a novel `tri-plane' representation of the radiance field which, in a memory efficient manner, factorises the latter into a triplet of 2D feature planes.
EG3D inspired several improvements such as GAUDI~\cite{bautista22gaudi:} and EpiGRAF~\cite{skorokhodov22epigraf:}.

Mesh-based 3D generators have been explored in~\cite{wu2020unsupervised}.
Recently, GET3D~\cite{gao2022get3d} replaced the radiance field with a signed distance function to regularise the representation of geometry.
The latter is converted into a mesh and rendered in a differentiable manner by using the marching tetrahedral representation~\cite{shen21deep}.

\paragraph{Modeling 3D with diffusion.}

Diffusion methods~\cite{sohl2015deep} have recently became the go-to framework for generative modeling of any kind, including 3D generative modeling.
The first applications of diffusion to 3D considered point-cloud generators trained on synthetic data~\cite{luo2021diffusion,zhou20213d,zeng2022lion}.

\paragraph{3D distillation of 2D diffusion models.}

More recently, DreamFusion~\cite{poole2022dreamfusion} ported the idea of distillation to diffusion models: they extract a neural radiance field so that its renders match the belief of a pre-trained 2D diffusion generator~\cite{saharia2022photorealistic,dhariwal2021diffusion,rombach2021stablediffusion}.
They introduce the Score Distillation Sampling (SDS) loss which makes distillation relatively efficient (but still in the order of several minutes for a single 3D sample).
Their generation can be conditioned by an image or by a textual description, making the process rather flexible.
Magic3D~\cite{lin22magic3d:} further increases the quality of the output by distilling a mesh-based 3D representation instead of a radiance field.

\paragraph{Image-conditioned 3D diffusion.}

The idea of distillation has been applied to few-view conditioned reconstruction in
\cite{watson22novel,gu2023nerfdiff,melas2023realfusion,zhou2022sparsefusion,deng2022nerdi}.
SparseFusion~\cite{zhou2022sparsefusion} employs a 3D-based new-view synthesis model~\cite{suhail2022generalizable} followed by a 2D diffusion upsampler.
They complete the process by 3D distillation, ensuring that the generated views of the object are consistent.
NeRFDiff~\cite{gu2023nerfdiff} and 3DiM~\cite{watson22novel} bypass an explicit 3D model and directly generate new views of an object using a 2D image generator and, in the case of NeRFDiff, refine the results using distillation.

While SparseFusion and NeRFDiff need to be trained on a dataset of object-centric multi-view images with pose information, RealFusion~\cite{melas2023realfusion} and NeRDi~\cite{deng2022nerdi} can be used for zero-shot monocular 3D reconstruction, starting from a pre-trained 2D diffusion model.
Given a single image as input, they automatically generate a prompt for the diffusion model, using a form of prompt inversion, and then use distillation to extract a radiance field.

\paragraph{Unconditional generation.}

Most relevantly to us, unconditional generation, \ie, generation which does not require either text or image conditioning, was explored in~\cite{wang2022rodin,shue20223d,muller22diffrf:,karnewar2023holodiffusion}.
While~\cite{muller22diffrf:,shue20223d,wang2022rodin} train generators given synthetic 3D ground truth, similar to us, HoloDiffusion~\cite{karnewar2023holodiffusion} is supervised only with real object-centric images and camera poses.
While HoloDiffusion was the first to demonstrate successful training on real image data, its renders contain considerably lower amount of detail than samples from a conventional 2D image generator that uses diffusion.
We thus leverage a 2D diffusion upsampler, conditioned on the lower-fidelity HoloDiffusion renders, to distill higher resolution images and, eventually, 3D models.

%% file: src/method.tex
\section{\name}%
\label{s:method}

We present \name, a method that can learn a high-quality diffusion-based 3D generator from a collection of multiview 2D images.
\name first obtains an unconditional low-resolution 3D sample using diffusion and then distills a high-resolution 3D radiance field representing a higher-quality version of the generated object.
We first summarize the Denoising Diffusion Probabilistic Models (DDPMs)~\cite{ho2020denoising} that we utilize in \cref{sec:ddpm}.
Then, we discuss the low-resolution 3D generator in \cref{sec:holo_diffusion}  followed by super-resolution distillation in \cref{sec:diff_guided_ft}.

\begin{figure*}
\centering
\includegraphics[width=\linewidth]{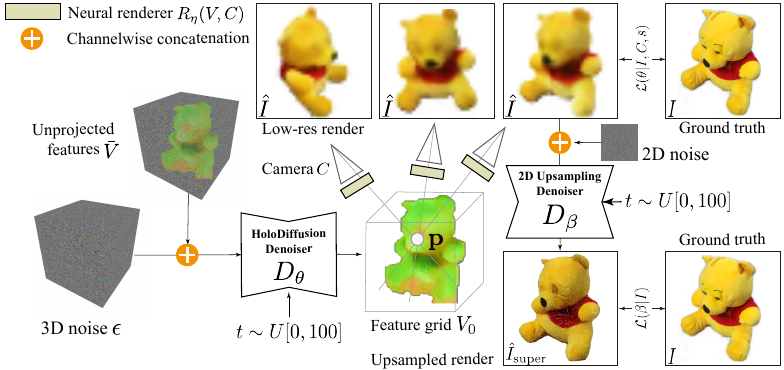}
\caption{
\textbf{Overview. }
\name, which trains the 3D denoiser network $D_\theta$, is augmented with the 2D `super-resolution' diffusion model $D_\beta$.
Both models are trained end-to-end by supervising their outputs with 2D photometric error.
\label{fig:method_part_1}
}
\end{figure*}


\subsection{Preliminaries: DDPMs}%
\label{sec:ddpm}

Let $x = x_0$ be a random vector whose probability distribution $p(x|y)$ we seek to model.
The DDPM~\cite{ho2020denoising} defines a hierarchy of latent variables $x_t$, $t=0,\dots,T$ and an
encoder $q$ 
comprising a sequence of Gaussian distributions
\begin{equation}\label{eq:q_sample_single}
q(x_t  | x_{t-1})
=
\mathcal{N}(x_t; \sqrt{\alpha_t}x_{t-1}, (1-\alpha_t)\mathbb{I}),
\end{equation}
where $\alpha_t, \dots, \alpha_T$ is a predefined `noising schedule'.
Given knowledge of $x_0$, a sample $x_t$ can be drawn in a closed-form directly from 
$
q(x_t | x_0)
=
\mathcal{N}(x_t; \sqrt{\bar\alpha_t}x_0, (1-\bar\alpha_t)\mathbb{I}),
$
where
$
\bar\alpha_t = \prod_{i=1}^{t} \alpha_i.
$
Hence, we can express $x_t$ as:
\begin{equation}
\label{eq:x_t_from_x_0}
x_t
=
\hat \epsilon_t(x_0)
=
\sqrt{\bar\alpha_t}x_0 + \sqrt{1 - \bar\alpha_t}\epsilon_t  
\end{equation}


The noising schedule is chosen such that $\bar \alpha_T \approx 0$.
In this manner, $q(x_T|x_0) \approx \mathcal{N}(x_T;0,\mathbb{I})$ is approximately normal, and so is $q(x_T)$.
To generate a sample $x = x_0$, we start by sampling $x_T$ from this normal distribution and then sample the intermediate latent variables backward.
This is done by using a variational approximation of the probabilities $q(x_{t-1}|x_t)$ given by the Gaussian factors:
\begin{equation}\label{eq:p_sample}
    p(x_{t-1} | x_{t}) = \mathcal{N}(x_{t-1}; \sqrt{\bar\alpha_{t-1}}D_\theta(x_t,t), (1 - \bar\alpha_{t-1})\mathbb{I})
\end{equation}
where $D_\theta$ is a neural network with parameters $\theta$.

The network $D_\theta$ is trained by maximizing the ELBO (Evidence Lower Bound), which reduces to the denoising objective~\cite{luo2022understanding}:
\begin{equation}\label{eq:denoising_objective}
\mathcal{L}(\theta)
=
\mathbb{E}_{t, \epsilon, x}
\left[
    \frac{\bar\alpha_{t-1}}{2(1-\bar\alpha_{t-1})^2}
    \|D_\theta(\hat \epsilon_t(x),t) - x\|^2
\right],
\end{equation}
where $t$ is sampled uniformly from $U[1,T]$.
Hence, $D_\theta(x_t,t)$ approximates the clean sample $x_0$ given the noisy sample $x_t$ (obtained using~\eqref{eq:x_t_from_x_0}).





\subsection{HoloDiffusion revisited}%
\label{sec:holo_diffusion}

Given a large dataset of 
3D models, the framework of \cref{sec:ddpm} could be used to train a corresponding probability distribution.
However, such a dataset is not available, and we must instead learn from 2D images of physical 3D objects.
Given a dataset containing several views of a large number of objects, we could use image-based reconstruction (\eg, using neural rendering) to obtain corresponding 3D models first, and then use those to train a diffusion model.
Instead, we adopt, and slightly upgrade, the HoloDiffusion method~\cite{karnewar2023holodiffusion}, which learns a 3D diffusion model \emph{directly} from the 2D images.

\paragraph{Training data.}

HoloDiffusion learns from a collection $\mathcal{D}$
of $N$ image sequences
$
s_i
=(I^i_j, C^i_j)_{j=1}^{N_{\text{frame}}},
$
$
i=1,\dots,N
$,
where frame $I^i_j \in \mathbb{R}^{3 \times H \times W}$ is an RGB image and
$
C^i_j \in \mathbb{R}^{4 \times 4}
$
is the corresponding camera projection matrix, collectively defining the motion of the camera.

\paragraph{3D representation and rendering.}

The shape and appearance of the object are represented by a voxel grid
$
V \in \mathbb{R}^{d \times S\times S\times S}
$
with resolution $S$ containing a $d$-dimensional feature vector per voxel.
Given a 3D point $\mathbf{p} \in \mathbb{R}^3$, its opacity $\sigma(\mathbf{p}) \in \mathbb{R}_+$ and color $c(\mathbf{p})\in\mathbb{R}_{[0, 1]}^3$ are obtained from the voxel grid by an MLP $M_\eta(V(\mathbf{p}))$ that takes as input the $d$-dimensional feature vector $V(\mathbf{p})$ extracted form the grid via trilinear interpolation~\cite{liu2020neural}.
The usual emission-absorption model~\cite{mildenhall20nerf:,DBLP:journals/tvcg/Max95a} is then used to implement a differentiable rendering function $R_\eta$, mapping the voxel grid $V$ and the camera viewpoint $C$ into an image
$
\hat I = R_\eta(V, C),
$
where $\eta$ are the parameters of the MLP\@.


\paragraph{Training scheme.}

HoloDiffusion leverages the DDPM framework (revised in the previous paragraphs) to recover the density $p(V)$ over voxel grids $x=V$ encoding plausible real-life objects.
In order to train a DDPM on such 3D data, we would need access to ground-truth 3D models $V$, which are not available.
HoloDiffusion addresses this problem by making three changes to DDPM\@.

First, it replaces the data denoising loss with a photometric reconstruction loss.
Given a pair $(I,C) \in s$ from one of the training sequences $s$, it replaces~\cref{eq:denoising_objective} with
$
\mathbb{E}_{t,\epsilon,C} \left[ \| I - R_\eta(D_\theta(\hat\epsilon_t(V), t), C) \|^2 \right]
$
where the goal is not to reconstruct the `clean' volume $V$ (which is unknown), but rather its image $I$ (which is known).

Second, also because the `clean' volume $V$ is not available, we cannot use \cref{eq:p_sample} to generate the noisy volumes $V_t$ to denoise;
the only exception is the last sample $V_T$, which is pure noise.
This suggests to adopt a `double denoising' step.
First, pure noise $V_T$ is fed into the denoiser to obtain an (approximate) version of $V_0 = D_\theta(V_T,T)$ of the clean volume $V_0 = V$.
Then, noise is applied to obtain
$
V_t
= \hat \epsilon_t(V_0)
= \sqrt{\bar\alpha_T} V_0  + \sqrt{1 - \bar\alpha_t} \epsilon_t
$
according to \cref{eq:p_sample}, and the latter is fed back into the denoiser as above.

Finally, there is the issue that unconditional generation of the clean volume $V_0$ from pure noise $V_T$ is difficult, especially in the first iterations of training.
On the other hand, the problem of \emph{view-conditioned} generation is considerably easier.
Hence, the third idea is to learn a \emph{conditional} generator, using a variable number of input views.
Specifically, given a training sequence $s$, the method extracts a random subset of frames $\bar s \subset s$ (which could be empty, which corresponds to unconditional generation).
Then, a feature volume
$
\bar V = \Phi(\bar s) \in \mathbb{R}^{d\times S\times S\times S}
$
is obtained from the selected frames.
This extracts 2D image features using a pre-trained and frozen 2D image encoder and then pools them in 3D via `unprojection'~\cite{kar2017learning,henzler21unsupervised} into $\bar V$, where $\bar V=0$ if $\bar s$ is empty.
Finally, these pooled features are used to condition the denoising network $V_0 = D_\theta(V_T,\bar V, T)$, which, on average, leads to a simpler reconstruction problem.

Putting it all together, the training loss becomes:
\begin{align}
\mathcal{L}(\theta | I, C, \bar s)
&=
\mathbb{E}_{t,\hat\epsilon,V_T} \left[ \| \hat I - I \|^2 \right],
\label{eq:photo_loss}
\\
\text{where}~~
\hat I &= R_\eta(D_\theta(V_t, \bar V, t),C), \label{eq:hatI} \\
V_t &= \hat \epsilon_t(V_0), \nonumber \\
V_0 &= D_\theta(V_T, \bar V, T), \nonumber \\
\bar V &= \Phi(\bar s).\nonumber
\end{align}
This loss is averaged over training sequences $s$, subsequences $\bar s \subset s$, and views $(I,C) \in s$ therein.
Note that this is slightly different than the original HoloDiffusion, where feature volume $\bar V$ and reconstructed volumes $V_t$ overlap as arguments of the denoiser; we found that keeping them separated in the formulation leads to more stable training and additionally allows for view-conditioned generation.


\subsection{\name}%
\label{sec:diff_guided_ft}

The method of~\cref{sec:holo_diffusion} learns to  generate 3D objects from 2D image supervision only, but the fidelity of the output is limited by the resolution at which the operations are carried out.
Increasing resolution is difficult due to the GPU memory impact of the voxel-based representation, so we seek a more efficient way to do so.
The idea is to incorporate a 2D super-resolution network (\cref{sec:2d_diff_post_proc}), trained end-to-end, that improves the output from the base model.
The super-resolved images are eventually fused back in an improved 3D model, which also has the benefit of further increasing view consistency (\cref{sec:fine_tuning_optim}).



\subsubsection{Integrating super-resolution}%
\label{sec:2d_diff_post_proc}

As shown in \cref{fig:method_part_1}, we augment the method of \cref{sec:holo_diffusion} with a lightweight refinement post-processor network that takes the 2D image $\hat I$ generated by the base model and outputs a higher quality version $\hat I_\text{super}$ of the same.
This can be thought of as a form of super-resolution;
however, due to the particular statistics of the input (`low-res') images $\hat I$ that HoloDiffusion generates, it is necessary to train this super-resolution network in an end-to-end fashion with HoloDiffusion, integrating the two models.

To make this integration seamless, we formulate super-resolution as another diffusion process that runs `in parallel' with 3D reconstruction.
Hence, the super-resolved image
$
\hat I_\text{super}
= D_\beta(I_t, \hat I, t)
$
is the output of a denoiser network (a lightweight U-Net), which takes as input the noised target image
$
I_t = \hat \epsilon_t(I)
$
and is also conditioned on the `low-res' output $\hat I = R_\eta(V,C)$ of HoloDiffusion from \cref{eq:hatI}.
This denoiser is trained with the DDPM loss:
\begin{align}
\mathcal{L}(\beta | I)
&=
\mathbb{E}_{t,\hat \epsilon} \left[ \| D_\beta(\hat \epsilon_t(I), \hat I, t) - I \|^2 \right].
\label{eq:photo_loss_super}
\end{align}

\paragraph{Training details}
The overall model ($D_\beta$ and $D_\theta$) is trained end-to-end by optimising the loss
$
\mathcal{L}(\theta|I,C,\bar s) + \mathcal{L}(\beta|I)
$
obtained by summing \cref{eq:photo_loss,eq:photo_loss_super}. 

As training data, we use a large dataset of images capturing object-centric scenes (\cite{reizenstein21common}).
In each training batch, we pick a random training scene $s$ and sample 15 different source images $\bar{s}_\text{src} \subset s$ which are unprojected to generate the feature volume conditioning $\bar{V}$.
Then, $\bar{V}$ is rendered into 4 random target views $\bar{s}_\text{tgt} \subset (s \setminus \bar{s}_\text{src})$ which allows to optimize the training image reconstruction loss
$
\mathcal{L}(\theta|I,C,\bar s) + \mathcal{L}(\beta|I)
$.
The latter uses the Adam optimizer with an initial learning rate of $5 \cdot 10^{-5}$ decaying tenfold whenever the loss plateaus until convergence.

\subsubsection{Fusing super-resolved views in 3D}%
\label{sec:fine_tuning_optim}

\begin{figure}
\centering
\includegraphics[width=\linewidth]{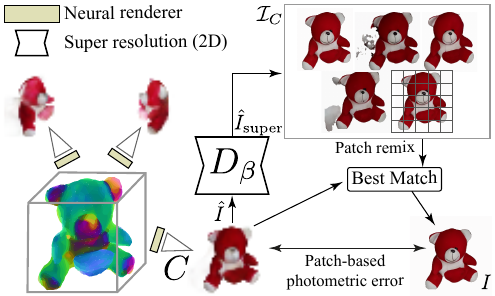}
\caption{\textbf{Distillation}.
\name distills a single high-resolution voxel grid $V^H_0$ by minimizing a top-k patch-remix loss $\mathcal{L}_\text{distil}$ between the grid renders $R_{\eta'}(V^H_0, C)$ and a bank $\mathcal{I}_C$ of $K=5$ high-res images output by the 2D diffusion upsampler $D_\beta$ for each scene camera $C$.}%
\label{fig:method_part_2}
\end{figure}

The method of \cref{sec:2d_diff_post_proc} leaves us with high-resolution views $\hat I_\text{super}$ of the generated 3D object.
However, we would like to obtain a single, high-quality 3D model, not just individual views of it.
In this section, we discuss how to take the super-resolved images and fuse them into such a model, while addressing the issues that these images are not perfectly view-consistent.

The basic idea is simple.
We can generate a certain number (\eg, 100) high-resolution images of the object from different viewpoints $C$ and then use a technique, akin to neural rendering, to fuse them back into a single 3D model.
However, there is a problem with this idea:
The model of \cref{sec:2d_diff_post_proc} generates high-quality views $I_\text{super}$, but these are \emph{view-dependent samples} from the distribution $p(\hat I_\text{super} | \hat I)$  where $\hat I = R_\eta(V,C)$ is the `low-res' output form HoloDiffusion.
Because super-resolving details is intrinsically ambiguous, there is no reason why samples $I_\text{super}$ taken from different viewpoints $C$ would be consistent (\cref{fig:variance_heatmap}).
Fusing them into a single 3D model would then result in a blurry appearance yet again.

As described in \cref{fig:method_part_2}, we address this issue in a principled manner by considering \emph{several} possible super-resolved images
$
\mathcal{I}_C = \{ I_\text{super} \sim p(\hat I_\text{super} | \hat I) \}
$
sampled from each given viewpoint $C$.
Then, we optimize a high-resolution voxel grid $V_0^H$ by minimizing the photometric loss:
\begin{equation}\label{e:min}
\mathcal{L}_\text{distil}(\eta', V_0^H | \mathcal{I}_C) = 
\mathbb{E}_C
    \left[
    \min_{I_\text{super} \in \mathcal{I}_C}
    \| I_\text{super} - R_{\eta'}(V_0^H, C) \|^2
    \right]
\end{equation}
where $R_{\eta'}(V_0^H, C)$ is the render of a high-resolution voxel grid $V_0^H \in \mathbb{R}^{d \times S' \times S' \times S'}, S' > S$ using the learnable renderer $R_{\eta'}$ with scene specific parameters $\eta'$.
Minimizing with respect to $I_\text{super}$ means that the 3D model must be consistent with at least \emph{one} of the possible super-resolved images, drawn from the distribution of super-resolved samples, for each view $C$.


\begin{figure}[!t]
\centering
\includegraphics[width=1.0\linewidth]{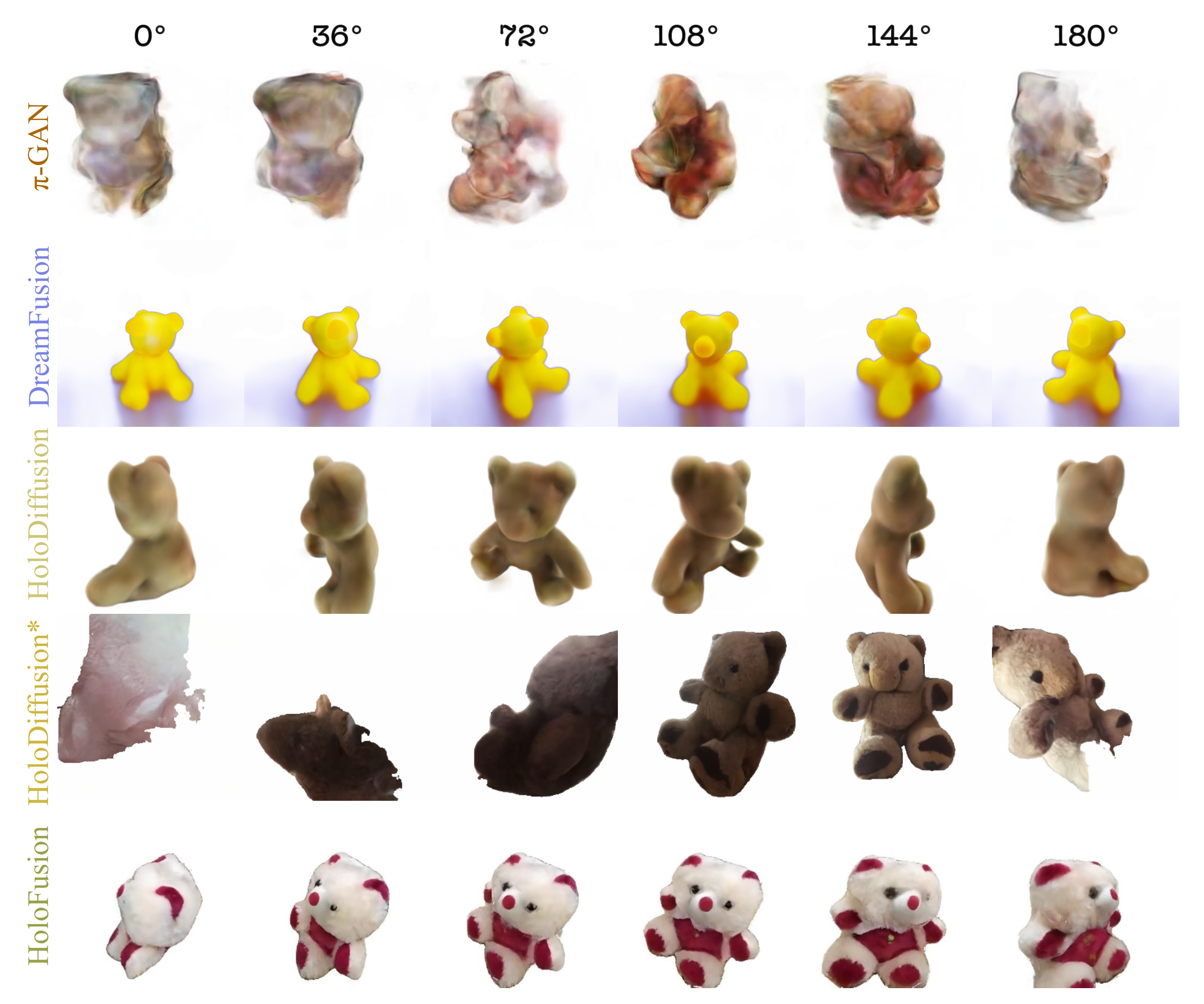}
\caption{
\textbf{Generated 3D samples visualized from a moving camera.}
\method{$\pi$-GAN} and \method{HoloDiffusion$^*$} fail to produce 3D view consistent samples, while \method{DreamFusion} suffers from the ``Janus'' problem (multiple heads). \label{fig:view_consistency}}
\end{figure}

\begin{figure}[b!]
\centering
\includegraphics[width=1.0\linewidth]{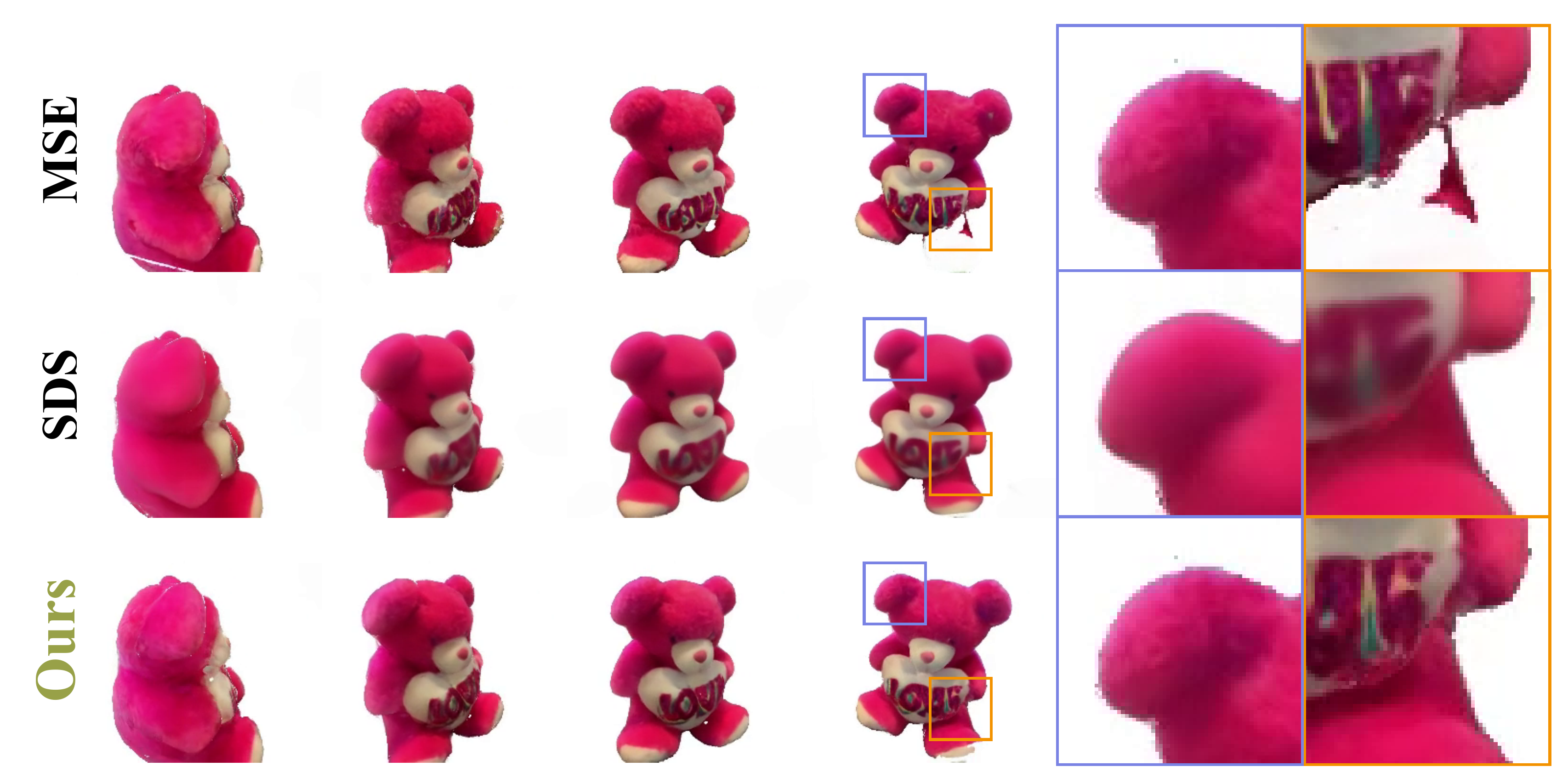}
\caption{
\textbf{Fusing views.} 
\method{Our} patch-remix (\cref{sec:fine_tuning_optim}) compared to the \textbf{SDS} and \textbf{MSE} distillation.
\methodb{MSE} has ``floaters'' and viewpoint inconsistencies, \methodb{SDS} over-smooths the texture.
\method{Ours} is robust and produces  superior quality.
}%
\label{fig:ablations}
\end{figure}

\paragraph{Patch remix.}
In practice, this approach requires a very large number of super resolved images $\mathcal{I}_C$ to be effective.
We found that we can significantly improve the statistical efficiency by performing the minimization at the level of individual patches.
Namely, we produce a stack of only $K=|\mathcal{I}_C|=5$ super resolved images and perform the minimization in \cref{e:min} at the level of small $16\times 16$ patches independently (effectively allowing super-resolved images to `remix' as needed to fit the generated view $R_{\eta'}(V_0^H, C)$).

\paragraph{Distillation details.}
$\mathcal{L}_\text{distil}$ is optimized independently for each generated scene with Adam (lr=$2 \cdot 10^{-4}$) for 25K steps until convergence.
While $\eta'$ is initialized using the pretrained multi-sequence weights $\eta$, $V_0^H$ is initialized by trilinearly upsampling the low-resolution volume $V_0$ output by HoloDiffusion.
Cameras $C$ are sampled at uniform azimuths with elevation fixed at object's equator.

%% file: src/experiments.tex
\section{Experiments}%
\label{s:experiments}

We begin with a description of the experiments conducted in \cref{ssec:exp_details}, followed by an analysis and discussion of the results in \cref{ssec:quant_and_qual}.

\subsection{Details}%
\label{ssec:exp_details}

\paragraph{Dataset.}

We experiment on the challenging large-scale Co3Dv2~\cite{reizenstein21common} dataset which is a popular choice for a real-world 3D reconstruction benchmark.
More specifically, $4$ categories are selected, \texttt{Apple}, \texttt{Hydrant}, \texttt{TeddyBear}, and \texttt{Donut}, with $500$ 3D-scenes per category for training.
Each 3D scene contains $\sim200$ images of the object of interest along with poses of their corresponding cameras.

\begin{figure*}[!t]
\vspace{0.5cm} 
\centering
\includegraphics[width=1.0\linewidth]{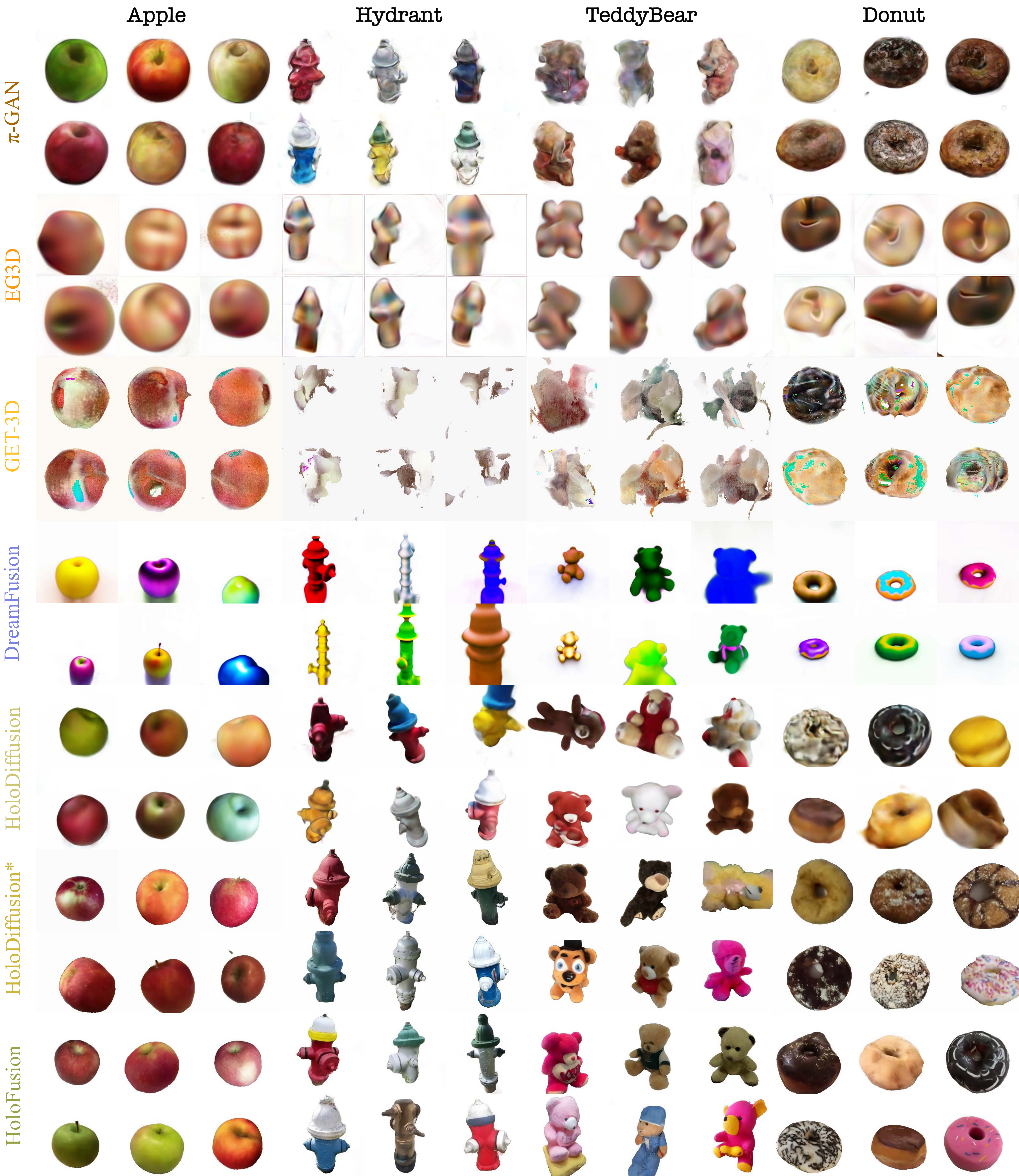}
\caption{3D samples generated by our \method{HoloFusion} compared to \method{$\pi$-GAN}, \method{EG3D}, \method{GET3D}, \method{HoloDiffusion}, \method{HoloDiffusion$^*$}, and the text-to-3D \method{Stable-DreamFusion}. \label{fig:mega_qualitative}}
\end{figure*}

\begin{table*}[t!]
\centering
\caption{FID ($\downarrow$) and KID ($\downarrow$) on 4 classes of Co3Dv2~\cite{reizenstein21common}.
We compare with 3D generative modeling baselines (rows 1--5); with an SDS distillation-based \method{Stable-DreamFusion} (row 6); and with ablations of our \method{HoloFusion} (rows 7--8). 
The column ``VP'' denotes whether renders of a method are 3D view-consistent or not.
}
\label{tab:base_qunat}
\resizebox{\linewidth}{!}{\input{src/tab_quant}}
\end{table*}

\paragraph{Baselines.}

We use two sets of baselines for comparison (\cref{tab:base_qunat}):
(i)~general 3D generative modeling baselines and
(ii)~diffusion distillation based baselines.
\method{$\pi$-GAN}~\cite{chan2021pi}, \method{EG3D}~\cite{chan2022efficient}, \method{GET3D}~\cite{gao2022get3d}, and \method{HoloDiffusion}~\cite{karnewar2023holodiffusion} are considered as the 3D generative baselines.
Along with \method{HoloDiffusion}, we also test the super-resolution integrated model (described in \cref{sec:2d_diff_post_proc}) \method{HoloDiffusion$^*$}.
For the distillation-based baselines, we consider the open-source implementation of \method{DreamFusion}~\cite{poole2022dreamfusion} titled \method{Stable-DreamFusion}~\cite{stable-dreamfusion}.
For the latter, scenes are generated by conditioning on prompts comprising names of Co3Dv2 categories extended with color and style modifier phrases leading to $\sim\!\!200$ prompts / 3D shapes per class.
More details regarding the prompt creation are in the supplementary.

\paragraph{Metrics.}

We use FID~\cite{heusel2017gans} and KID~\cite{binkowski2018demystifying} to compare the quality of our 2D renders, as these are commonly used to assess 2D and 3D generators.


\subsection{Quantitative and qualitative analysis}%
\label{ssec:quant_and_qual}

\cref{tab:base_qunat} evaluates quantitatively while \cref{fig:mega_qualitative} qualitatively.
Furthermore, \cref{fig:view_consistency} compares rendering view-consistency.

\method{HoloFusion (Ours)} yields better FID/KID scores than the general 3D generative baselines except for \method{$\pi$-GAN} on \texttt{Apple} and \texttt{Donut} classes.
However, since \method{$\pi$-GAN} does not guarantee view consistency by design, it essentially acts as a 2D image GAN, and thus does better on the 2D FID/KID metrics, but it generates significantly view-inconsistent renders (see \cref{fig:view_consistency} and the supplementary).

We observed that the other 3D-GAN baselines, \method{EG3D} and \method{GET3D}, are prone to collapsing to a single adversarial sample leading to poor FID/KID scores.
The latter is probably due to the 3D misalignment of the CO3Dv2 sequences across instances, which makes training harder.

\begin{figure}[t!]
\centering
\includegraphics[width=1.0\linewidth]{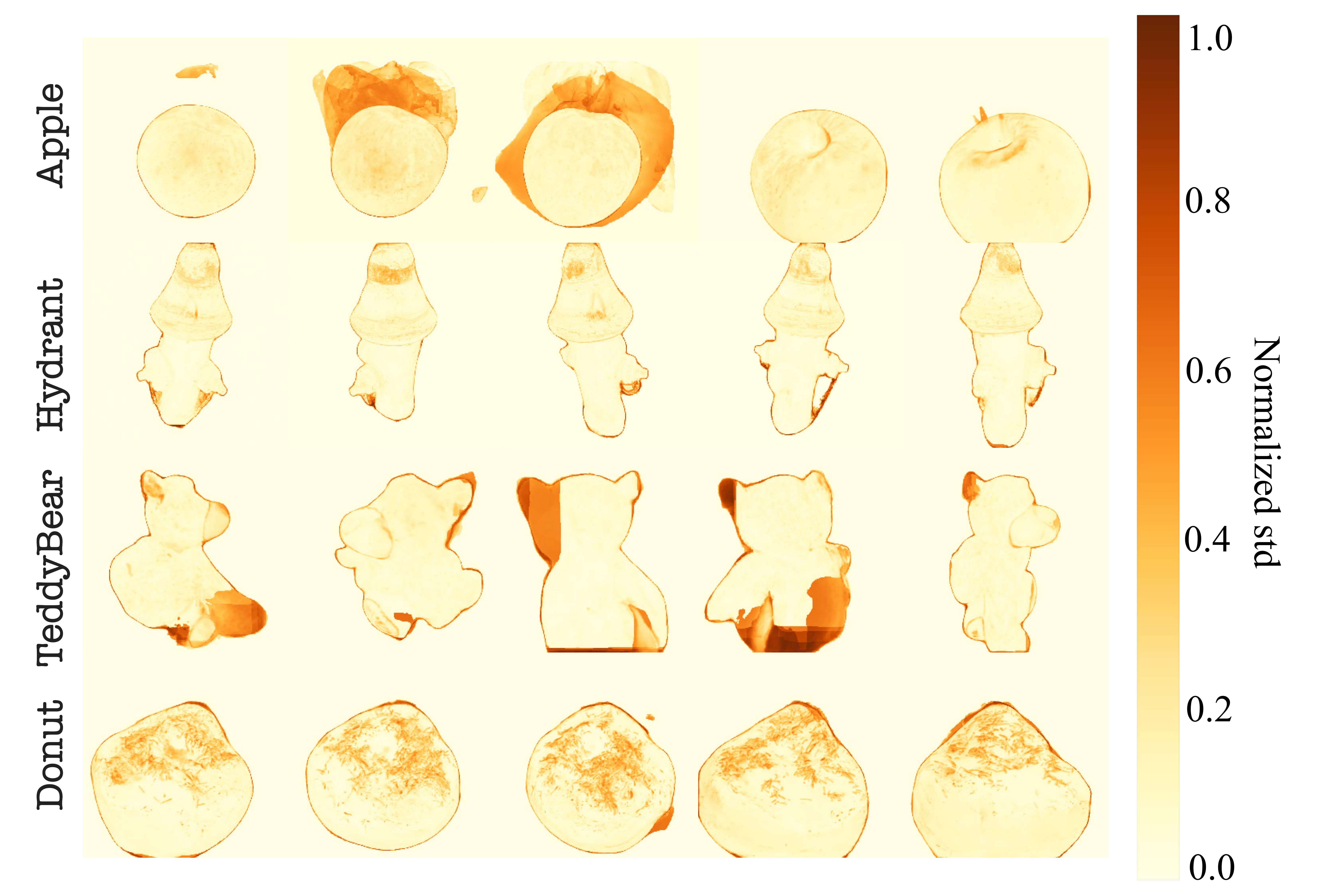}
\caption{Heatmaps illustrating the per-pixel color variance of $K=10$ hypothesis produced by the upsampler $D_\beta$.
Some samples contain artifacts around the object boundaries which correspond to the high-variance regions in the figure.
Our top-K patch-remix increases robustness by allowing the loss to discard such artifacts during distillation.}%
\label{fig:variance_heatmap}
\end{figure}

\method{HoloFusion} also outperforms the text-to-3D \method{Stable-DreamFusion} on both FID/KID.
\method{Stable-DreamFusion} yields good shapes, but produces synthetic-looking and overly-smooth textures and thus performs poorly when compared to the real-world images of Co3Dv2.
As evident from the \texttt{TeddyBear} samples, the method also suffers from the ``Janus'' issue.

Compared to \method{HoloDiffusion}, we improve the FID/KID scores by a significant margin, mainly due to the more photo-realistic renders that include high-frequency details.
Although the 2D Diffusion upsampler of \method{HoloDiffusion$^*$} produces renders with the highest amount of details yielding scores similar to ours, they are not 3D view-consistent (as apparent from \cref{fig:view_consistency} and as explained in \cref{sec:diff_guided_ft}).

\paragraph{Ablations.}
In \cref{tab:base_qunat} and in \cref{fig:ablations} we ablate components of our \method{HoloFusion} to verify their contribution.

The first variant, \method{HoloFusion (SDS)}, replaces the Top-k patch-remixed distillation loss with the score distillation sampling (SDS) gradient as proposed in \cite{poole2022dreamfusion}.
As apparent from \cref{fig:ablations} and from the lower scores, SDS washes out all the high-frequency details in the textures.

Secondly, \method{HoloFusion (MSE)} reduces the number of upsampling hypotheses $\mathcal{I}$ to the minimum of $|\mathcal{I}|=1$.
Even though this slightly improves the 2D metrics, as can be seen from \cref{fig:ablations}, the samples lack view-consistency and introduce ``floaters''.
In \cref{fig:variance_heatmap} we further illustrate the variability of the upsampling hypotheses.

%% file: src/tab_quant.tex
\begin{tabular}{ l  c  rr  rr  rr  rr  rr }
\toprule\small
 method & VP    & \multicolumn2c{\texttt{Apple}}        & \multicolumn2c{\texttt{Hydrant}} &
                  \multicolumn2c{\texttt{TeddyBear}}    & \multicolumn2c{\texttt{Donut}}   &
                  \multicolumn2c{Mean} \\
                  \cmidrule(lr){3-4}      \cmidrule(lr){5-6}     \cmidrule(lr){7-8}     \cmidrule(lr){9-10}
                  \cmidrule(lr){11-12}
        &       & \multicolumn1c{\footnotesize{FID $\downarrow$}} & \multicolumn1c{\footnotesize{KID $\downarrow$}} &
                  \multicolumn1c{\footnotesize{FID $\downarrow$}} & \multicolumn1c{\footnotesize{KID $\downarrow$}} &
                  \multicolumn1c{\footnotesize{FID $\downarrow$}} & \multicolumn1c{\footnotesize{KID $\downarrow$}} &
                  \multicolumn1c{\footnotesize{FID $\downarrow$}} & \multicolumn1c{\footnotesize{KID $\downarrow$}} &
                  \multicolumn1c{\footnotesize{FID $\downarrow$}} & \multicolumn1c{\footnotesize{KID $\downarrow$}} \\
\midrule

\method{$\pi$-GAN}~\cite{chan2021pi}                     &  \xmark & 49.3  & 0.042 & 92.1  & 0.080 & 125.8  & 0.118 & 99.4  & 0.069
                                                          & 91.7 & 0.077 \\

\method{EG3D}~\cite{chan2022efficient}                   &  \cmark & 170.5 & 0.203 & 229.5 & 0.253 & 236.1 & 0.239 & 222.3 & 0.237
                                                          & 214.6 & 0.233 \\

\method{GET3D}~\cite{gao2022get3d}                       &  \cmark & 179.1 & 0.190 & 303.3 & 0.380 & 244.5 & 0.280 & 209.9 & 0.230
                                                          & 234.2 & 0.270 \\

\method{HoloDiffusion}~\cite{karnewar2023holodiffusion}  &  \cmark & 94.5     & 0.095 & 100.5 & 0.079 & 109.2  & 0.106 & 115.4 & 0.085
                                                          & 122.5  & 0.102 \\

\method{HoloDiffusion$^*$}                        &  \xmark & 55.9   & 0.045 & 62.6 & 0.045 & 116.6 & 0.101 & 99.6 & 0.079
                                                          &  83.7 & 0.068  \\

\midrule

\method{Stable-DreamFusion}~\cite{stable-dreamfusion}    &  \cmark & 139.0  & 0.104 & 185.2  & 0.132 & 183.4 & 0.125 & 169.3 & 0.114
                                                          &  169.2 & 0.119 \\

\midrule

\method{HoloFusion (MSE)}                                 &  \xmark & 72.7   & 0.067 & 62.2 & 0.045 & 87.2  & 0.076 & 109.0 & 0.099
                                                           & 82.8  & 0.072  \\
\method{HoloFusion (SDS)}                                 &  \cmark & 123.0  & 0.105 & 77.1 & 0.058 & 117.8 & 0.090 & 142.8 & 0.087
                                                           & 115.2 & 0.085  \\

\midrule

\methodb{HoloFusion (Ours)}                      &  \cmark & 69.2  & 0.063 & 66.8 & 0.047   & 87.6 & 0.075 & 109.7 & 0.098
                                                          & 83.3 & 0.071 \\

\bottomrule

\end{tabular}

%% file: src/outro.tex
\section{Conclusion}%
\label{s:conclusions}

We have presented a hybrid diffusion-based method that can generate high-quality 3D neural radiance fields of real-life object categories.
Our method starts by producing coarse 3D models whose renders are independently super-resolved, and finally consolidated using a robust distillation process.
We evaluated our method on the Co3D v2 dataset and presented 3D-consistent, diverse, and high-quality results superior to all competing baselines.

Our method suffers from limitations that can be addressed in future work.
First, our method is slow to sample from as the sampling process takes about 30 mins for each generation, because it is still a distillation-based method.
An interesting extension would be to train another network to directly distill a set of super-resolved images, without requiring explicit optimization during inference.
Second, we do not produce an explicit surface representation (\eg, a mesh), which could be done by integrating a differentiable mesh render in the loop as done in some prior work.

%% file: src/acknowledgements.tex
\section{Acknowledgements}

Animesh and Niloy were partially funded by the European Union’s Horizon 2020 research and innovation programme under the Marie Skłodowska-Curie grant agreement No.~956585. This research has been partly supported by MetaAI and the UCL AI Centre.
Finally Animesh is grateful to \href{http://www.rabinezra.info/}{The Rabin Ezra Scholarship Fund} being a recipient of their esteemed fellowship for the year 2023.